\documentclass[lettersize,journal]{IEEEtran}
\usepackage{amsmath,amsfonts}
\usepackage{algorithm}
\usepackage{algorithmic}
\usepackage{array}
\usepackage[caption=false,font=normalsize,labelfont=sf,textfont=sf]{subfig}
\usepackage{textcomp}
\usepackage{stfloats}
\usepackage{url}
\usepackage{verbatim}
\usepackage{graphicx}
\def\BibTeX{{\rm B\kern-.05em{\sc i\kern-.025em b}\kern-.08em
    T\kern-.1667em\lower.7ex\hbox{E}\kern-.125emX}}
\usepackage{balance}

\usepackage{multicol}
\usepackage{multirow}
\usepackage{amsthm}
\usepackage{booktabs}
\usepackage{xcolor}   
\usepackage{amssymb}  

\usepackage{hyperref}

\begin{document}

\title{\textit{DanceText}: A Training-Free Layered Framework for Controllable Multilingual Text Transformation in Images}

\author{Zhenyu Yu$^{1}$, Mohd Yamani Idna Idris$^{1}$, Hua Wang$^{2,*}$, Pei Wang$^{3,*}$, Rizwan Qureshi$^{4}$, Shaina Raza$^{5}$, Aman Chadha$^{6}$, Yong Xiang$^{7}$, Zhixiang Chen$^{8}$ \thanks{
$^{1}$ Universiti Malaya, Kuala Lumpur 50603, Malaysia.\\
$^{2}$ Zhejiang University of Finance and Economics Dongfang College, Jiaxing 314408, China.\\
$^{3}$ Kunming University of Science and Technology, Kunming 650500, China.\\
$^{4}$ University of Central Florida, Orlando, FL 32816, USA.\\
$^{5}$ Toronto metropolitan university, Toronto, ON M5B 2K3, Canada. Workdone outside.
$^{6}$ Amazon Web Services, Washington, 98109, USA. Workdone outside Amazon.\\
$^{7}$ Deakin University, Burwood VIC 3125, Australia.\\
$^{8}$ University of Sheffield, Sheffield S10 2TN, United Kingdom.\\
Corresponding email: yuzhenyuyxl@foxmail.com\\
}}

\markboth{Journal of \LaTeX\ Class Files,~Vol.~18, No.~9, September~2020}%
{\textit{DanceText}: A Training-Free Layered Framework for Controllable Multilingual Text Transformation in Images}

\maketitle

\begin{abstract}
    We present \textit{DanceText}, a training-free framework for multilingual text editing in images, designed to support complex geometric transformations and achieve seamless foreground-background integration. While diffusion-based generative models have shown promise in text-guided image synthesis, they often lack controllability and fail to preserve layout consistency under non-trivial manipulations such as rotation, translation, scaling, and warping. To address these limitations, \textit{DanceText} introduces a \textbf{layered editing strategy} that separates text from the background, allowing geometric transformations to be performed in a modular and controllable manner. A \textbf{depth-aware module} is further proposed to align appearance and perspective between the transformed text and the reconstructed background, enhancing photorealism and spatial consistency. Importantly, \textit{DanceText} adopts a fully \textbf{training-free} design by integrating pretrained modules, allowing flexible deployment without task-specific fine-tuning. Extensive experiments on the AnyWord-3M benchmark demonstrate that our method achieves superior performance in visual quality, especially under large-scale and complex transformation scenarios. Code is avaible at https://github.com/YuZhenyuLindy/DanceText.git.
\end{abstract}



\begin{figure*}[!ht]
    \centering
    \includegraphics[width=0.8\linewidth]{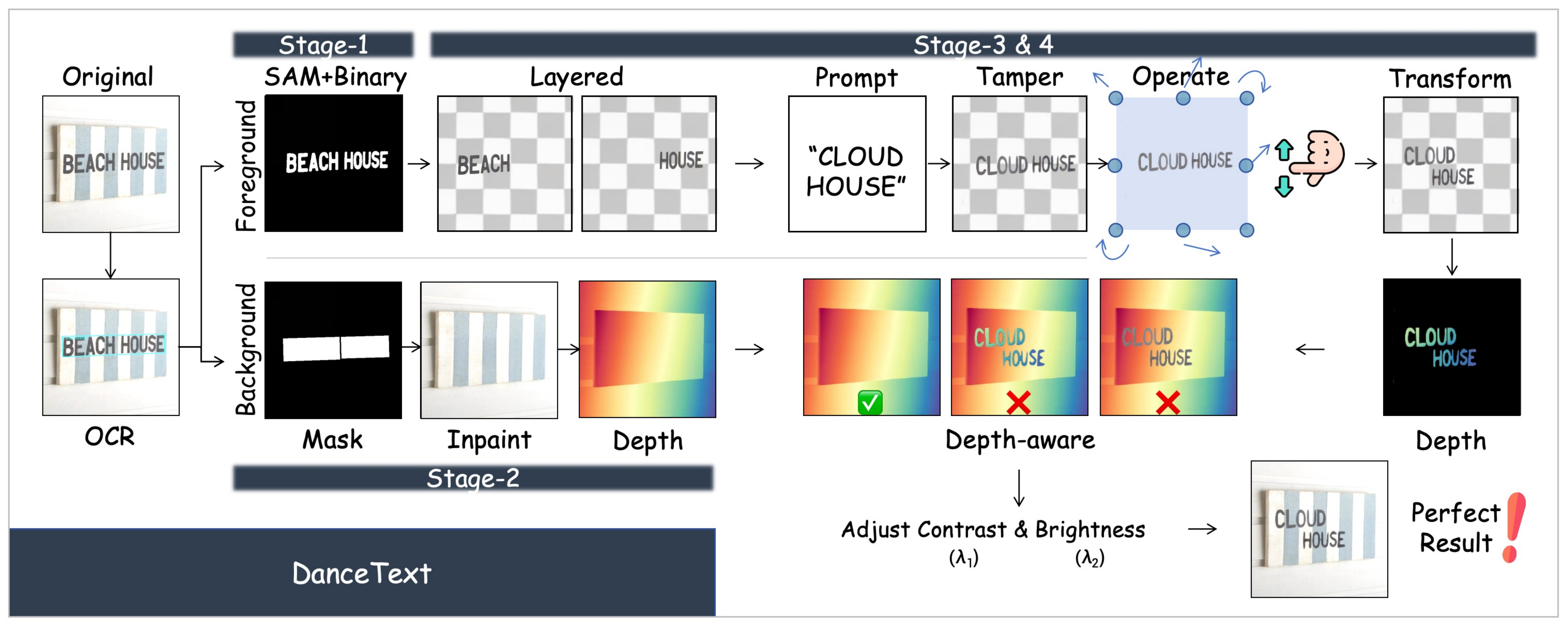}
    \caption{Overview of the \textit{DanceText} framework. The pipeline consists of four stages: (1) OCR- and SAM-based foreground extraction; (2) Background inpainting after text removal; (3) Geometric transformation of the text layer (rotation, translation, scaling, warping); (4) Depth-aware composition to ensure realistic integration.}
    \label{fig_method_overview}
\end{figure*}

\section{Introduction}

With the rapid development of generative artificial intelligence, diffusion models have emerged as a core paradigm in multimodal research~\cite{yu2025yuan,yu2025qrs,yu2024improved}. These models have demonstrated impressive capabilities in tasks such as semantic-guided image generation, vision-language alignment, and visual content enhancement~\cite{ramesh2022hierarchical,saharia2022photorealistic}. In this context, multilingual text generation and fine-grained editing have become key challenges for enhancing visual expressiveness and enabling cross-lingual communication~\cite{rombach2022high,balaji2022ediffi,10688076,10687425}. For example, AnyText~\cite{tuo2024anytext} combines diffusion models with OCR modules to support controllable multilingual text rendering and style transfer, achieving notable performance in multilingual scene text generation. However, existing methods primarily focus on text content generation and appearance adaptation, lacking explicit modeling of the spatial and geometric attributes of text. Consequently, they struggle to maintain layout consistency and structural coherence under complex transformations such as rotation, translation, scaling, and warping.

From a modeling perspective, high-quality text editing requires not only semantic consistency and visual fidelity, but also fine-grained geometric controllability and spatial coherence with the background. This process involves foreground-background disentanglement, modular manipulation of editable text regions, and depth-consistent re-composition. Most existing methods adopt end-to-end diffusion pipelines, where the generation is globally noise-driven and lacks region-specific controllability. Furthermore, the absence of modular decomposition and explicit geometric reasoning severely limits their adaptability and generalization in complex visual scenarios~\cite{brooks2022instructpix2pix,chen2023textdiffuser, qureshi2025thinking}.

To address these challenges, we propose \textit{DanceText}, a \textbf{training-free} framework for multilingual text editing and transformation in images. Unlike end-to-end generative systems, \textit{DanceText} is constructed entirely from pretrained modules in a modular fashion, leveraging OCR for text detection, SAM for region segmentation, and LaMa-based diffusion inpainting for background reconstruction. This design eliminates the need for task-specific training or fine-tuning while preserving high quality editing flexibility. At its core, \textit{DanceText} adopts a \textbf{layered editing strategy} that explicitly separates text from background content, enabling controllable geometric transformations such as rotation, translation, scaling, and warping. In addition, we propose a \textbf{depth-aware module} to ensure photometric and geometric consistency between the transformed text and the restored background. Extensive experiments show that \textit{DanceText} achieves superior visual quality under complex transformation scenarios, demonstrating its effectiveness and generalizability in multilingual text editing tasks.

The main \textbf{contributions} of this work are summarized as follows:
\begin{itemize}
    \item \textbf{Layered text transformation.} We propose a layered editing framework that explicitly decouples text from background regions, enabling controllable geometric transformations (e.g., rotation, translation, scaling, warping) in a modular and controllable manner.
    
    \item \textbf{Depth-aware module.} We propose a depth-aware composition module that explicitly models photometric and spatial coherence between the transformed text and the reconstructed background, thereby enabling seamless integration and enhancing overall visual consistency.

    \item \textbf{Training-free architecture.} \textit{DanceText} is entirely training-free, leveraging pretrained modules without task-specific fine-tuning, making the framework efficient, generalizable, and deployment-friendly across diverse editing scenarios.
\end{itemize}

\section{Related Work}

\subsection{Image Generation and Editing with Diffusion Models}

Diffusion models have emerged as a foundational paradigm in generative artificial intelligence, achieving state-of-the-art performance in text-to-image synthesis~\cite{ramesh2022hierarchical, rombach2022high}. Models such as DALL-E 2 and Stable Diffusion leverage large-scale text-image pretraining to generate semantically aligned and visually realistic content~\cite{huang2025diffusion}. These approaches have spurred substantial progress in vision-language modeling, inspiring further research into fine-grained content control and guided image manipulation~\cite{saharia2022photorealistic,yu2023superpixel}.

Recent efforts have extended diffusion models toward the task of text generation and editing within images. Representative methods include TextDiffuser~\cite{chen2023textdiffuser}, which employs character-level segmentation to facilitate controllable text synthesis, and GlyphControl~\cite{yang2023glyphcontrol}, which introduces glyph-conditioned diffusion processes to improve style and layout fidelity. AnyText~\cite{tuo2024anytext} further advances this line of research by supporting multilingual text rendering through a combination of OCR-guided positioning and diffusion-based generation.

While these approaches demonstrate impressive results in static text generation, they remain limited in their capacity to support spatial transformations or dynamic text rearrangement. In particular, operations such as rotation, scaling, and repositioning are not natively supported by existing diffusion-based pipelines, highlighting a gap in controllable and geometry-aware text editing—a gap that our work aims to address.

\subsection{Controllable Text Editing and Spatial Transformation}

Precise control over text placement and transformation is essential for diffusion-based scene text editing. Existing methods primarily support spatial specification during the initial generation phase. For instance, TextDiffuser~\cite{chen2023textdiffuser} leverages segmentation masks to constrain text position, while GlyphControl~\cite{yang2023glyphcontrol} introduces explicit positional markers to enhance localization and layout fidelity. However, these approaches typically lack support for post-generation manipulation, offering limited controllability in adjusting existing text spatially.

InstructPix2Pix~\cite{brooks2022instructpix2pix} enables region-based image editing via natural language prompts, and has proven effective for general appearance modifications. Nevertheless, it does not incorporate dedicated mechanisms for geometry-aware text relocation or layout adjustment, especially under non-rigid transformations.

Overall, current methods remain constrained in their capacity to handle dynamic spatial reconfiguration of text elements, particularly under geometric operations such as rotation, translation, scaling, or warping. Addressing these limitations requires a controllable framework that decouples content manipulation from spatial control, and our proposed method fills this gap.

\subsection{Foreground-Background Disentanglement}

Foreground-background separation has long been a fundamental strategy in image editing, particularly useful in scenarios involving cluttered or semantically rich scenes. Traditional methods such as Content-Aware Fill~\cite{barnes2009patchmatch} perform patch-based feature matching to reconstruct occluded regions. While effective for small-scale edits, these approaches often fail in large or structurally complex areas, resulting in blurry or inconsistent completions.

The emergence of deep generative inpainting models has significantly improved reconstruction quality. DeepFill~\cite{yu2018generative} models long-range dependencies via contextual attention, and EdgeConnect~\cite{nazeri2019edgeconnect} introduces edge-guided priors for structure-preserving synthesis. LaMa~\cite{suvorov2022resolution} further advances this line of work by employing Fourier convolutions and multi-scale feature learning to achieve high-quality inpainting on large missing regions. However, most of these frameworks are optimized for generic object removal and do not account for the unique geometric and semantic characteristics of text.

To address this gap, we propose a task-specific disentanglement pipeline that combines OCR-guided foreground extraction with background inpainting. This layered design enables precise separation of text from the background, allowing flexible geometric transformations, while maintaining global scene consistency.

\begin{figure*}[!ht]
    \centering
    \includegraphics[width=0.8\linewidth]{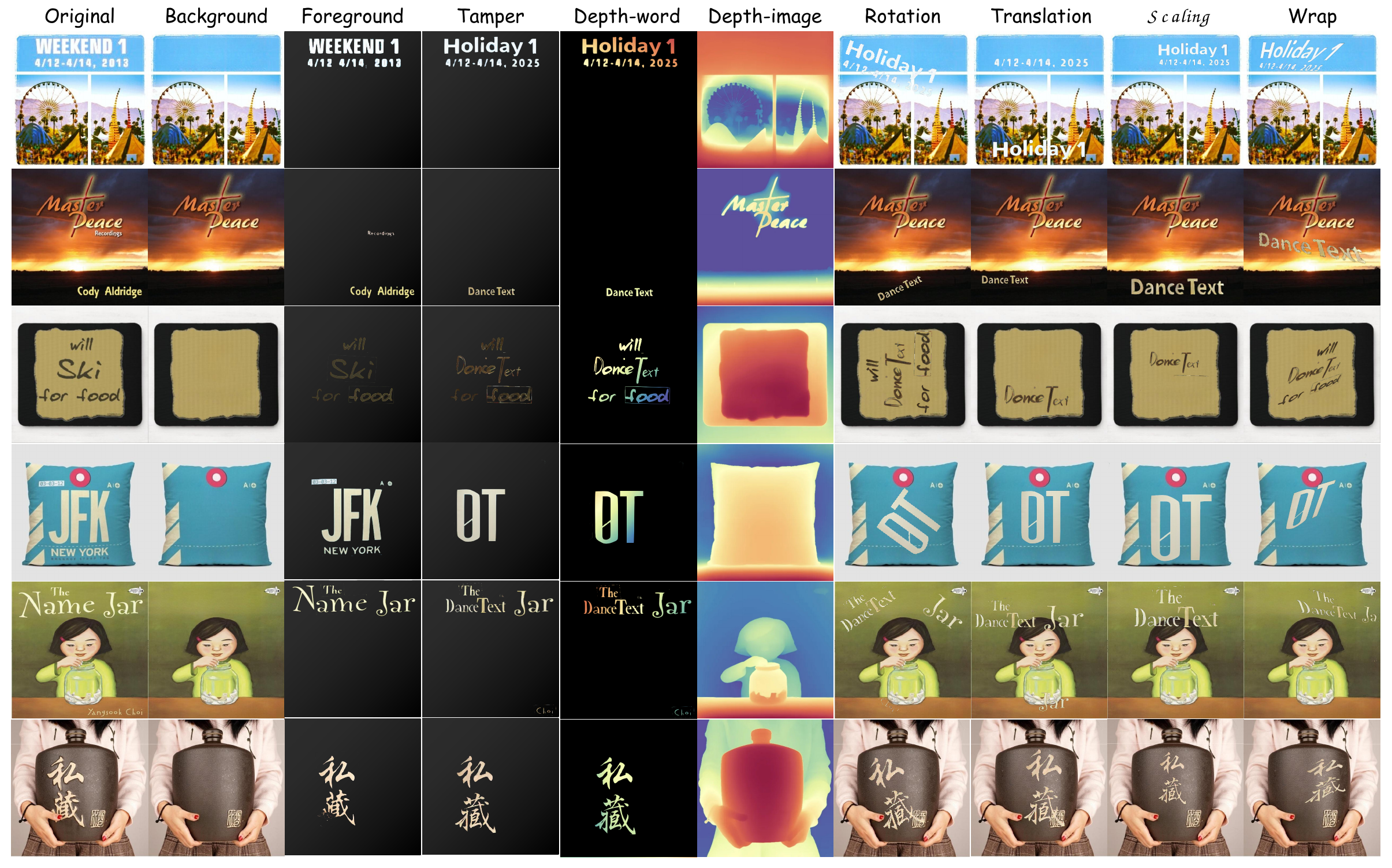}
    \caption{Visualization of the \textit{DanceText} pipeline. The process consists of the following stages: (1) \textbf{Original} input image, (2) \textbf{Background} extraction, (3) \textbf{Foreground} text separation, (4) \textbf{Tamper} for text modifications, (5) \textbf{Depth-word} for text depth processing, (6) \textbf{Depth-image} for global depth refinement, (7) \textbf{Rotation}, (8) \textbf{Translation}, (9) \textbf{Scaling}, and (10) \textbf{Wrap}.}
    \label{fig_result_dancetext}
\end{figure*}

\begin{figure*}[!ht]
    \centering
    \includegraphics[width=0.8\linewidth]{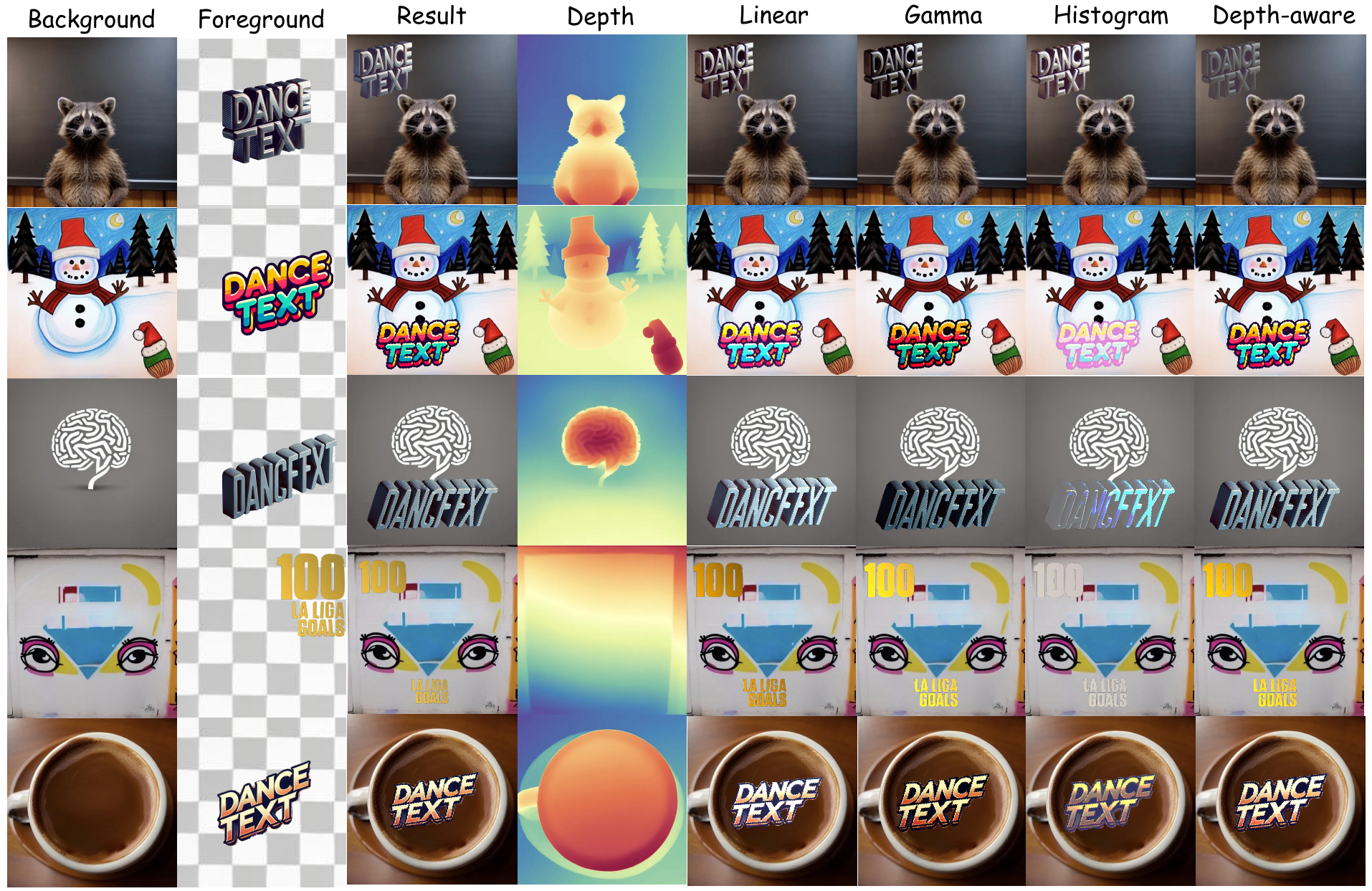}
    \caption{Comparison of foreground-background composition methods, including {Linear}, {Gamma}, {Histogram matching}, and {Depth-aware}.}
    \label{fig_compare_composition}
\end{figure*}

\section{Method}

\subsection{Overview}

We present \textit{DanceText}, a modular and training-free framework for multilingual text editing in images (Figure~\ref{fig_method_overview}). The proposed design facilitates fine-grained control over geometric transformations while mitigating background interference, thereby enabling seamless integration within complex visual scenes. The overall workflow comprises the following stages:

\textbf{Stage 1: Foreground Extraction.} EasyOCR detects text regions, SAM segments glyphs, and $k$-Means binarization removes edge background noise to form text layers.

\textbf{Stage 2: Background Restoration.} The text-occluded regions are removed and restored using the LaMa inpainting model to recover a clean background.

\textbf{Stage 3: Text Tampering and Transformation.} The text is edited based on prompts and undergoes user-controllable geometric transformations to adjust its position and shape according to user intent.

\textbf{Stage 4: Depth-Aware Composition.} Depth-guided modulation is performed to align appearance attributes, enabling visually coherent integration of foreground and background layers.

\subsection{Foreground Extraction}

The initial stage of \textit{DanceText} aims to isolate text regions as a structured foreground representation for subsequent transformation. This module comprises three components: OCR-guided localization, glyph-level segmentation, and $k$-Means-based refinement.

\textbf{Text detection and mask generation.}  
Given an input image $I \in \mathbb{R}^{H \times W \times 3}$, the EasyOCR model~\cite{easyocr2020} $\mathcal{D}$ is employed to detect text regions and recognize their content, yielding a set of tuples:

\begin{equation}
    \{(B_i, T_i)\}_{i=1}^N = \mathcal{D}(I),
\end{equation}
where each $B_i = (x_i, y_i, w_i, h_i)$ denotes the bounding box of the $i$-th text region, and $T_i$ is the corresponding recognized text. The set of bounding boxes $\{B_i\}$ is further used to construct an initial binary mask $M$, which marks the candidate foreground areas for subsequent segmentation and refinement.

\textbf{Glyph segmentation and clustering refinement.}  
To obtain a more precise delineation of text shapes, the Segment Anything Model (SAM)~\cite{kirillov2023segment} is applied to extract glyph-level masks $\hat{M}$ within the regions indicated by $M$. While SAM provides high-quality segmentation, it may occasionally include non-text elements due to low contrast or background clutter. To mitigate this, a clustering-based refinement is introduced.

Specifically, each segmented region is processed by applying $k$-Means clustering ($k=2$) to RGB pixel values. The cluster with the greater number of pixels is selected as the foreground, while the smaller one is discarded as noise. This filtering strategy helps eliminate background interference along glyph boundaries.

\textbf{Foreground layer construction.}  
The refined mask $\hat{M}$ is applied to the input image to generate the final foreground layer:

\begin{equation}
I_{\text{fg}} = I \odot \hat{M},
\end{equation}
where $\odot$ denotes element-wise multiplication. The resulting $I_{\text{fg}}$ contains only text glyphs, enabling decoupled transformation and reintegration in subsequent stages.

\subsection{Background Restoration}

Following foreground extraction, the original background—previously occluded by text—must be reconstructed to enable realistic reintegration in subsequent stages. To this end, we adopt LaMa~\cite{suvorov2022resolution}, an inpainting model capable of handling large masked regions while maintaining perceptual coherence.

Given the input image $I$ and the binary text mask $M$ obtained from the previous stage, the restored background $\hat{I}_{\text{bg}}$ is generated by applying the LaMa inpainting:

\begin{equation}
\hat{I}_{\text{bg}} = \mathcal{L}(I, M),
\end{equation}
where $\mathcal{L}$ denotes the inpainting process. By leveraging frequency-domain convolutions and dilated architectures, LaMa synthesizes plausible textures and structures in occluded regions, producing a clean, text-free background with minimal perceptual artifacts.

\subsection{Text Tampering and Transformation}

After foreground-background separation, the extracted text can be semantically tampered and spatially transformed to fit target visual designs. This stage comprises two operations: text tampering and geometric transformation.

\textbf{Text tampering.}  
We define text tampering as a prompt-guided content modification process. Given the original text content $T_i$ from Stage~1 and a user-specified prompt $S_i$, a text editing model $\mathcal{T}$ (AnyText~\cite{tuo2024anytext}) generates the modified output:

\begin{equation}
T'_i = \mathcal{T}(S_i, T_i),
\end{equation}
where $T'_i$ denotes the tampered version with revised semantics. The updated text is rendered with font and style consistent with the original image, preserving visual coherence.

\textbf{Geometric transformation.}  
To support user-driven spatial adjustment, a series of geometric transformations are applied to the foreground layer $I_{\text{fg}}$. The supported operations include:  
\textbf{Rotation} alters the orientation to match the perceived scene geometry;  
\textbf{Translation} repositions the text to a user-specified location;  
\textbf{Scaling} adjusts the text size to achieve visual balance;  
\textbf{Warping} deforms the text shape to conform to curved or irregular surfaces.  
These transformations enable fine-grained and user-controllable adaptation.

\subsection{Depth-Aware Composition}

In the final stage of the pipeline, the geometrically transformed text must be seamlessly integrated with the restored background. A key challenge in this step lies in achieving perceptual consistency.

\textbf{Depth-Aware Adjustment.}  
Directly overlaying the transformed foreground onto the reconstructed background often results in perceptual inconsistencies, such as unnatural lighting, poor shading continuity, and visual misalignment with scene geometry. These issues arise because the inserted text lacks proper adaptation to the underlying depth and illumination context of the scene.

To mitigate this, we propose a \textbf{depth-aware adjustment module} that modulates the foreground appearance based on estimated depth information from the background. Specifically, we first estimate a depth map \( D \in \mathbb{R}^{H \times W} \) from the inpainted background \( \hat{I}_{\text{bg}} \) (Stage~2). We then generate an initial depth map \( D_{\text{fg}} \) for the transformed foreground based on its placement. The goal is to refine \( D_{\text{fg}} \) so that it aligns with the surrounding background depth at the composition boundary.

Let \( \Delta D = D - D_{\text{fg}} \) denote the pixel-wise depth difference. We define an adjustment function \( \mathcal{F} \) that modifies the foreground appearance according to \( \Delta D \), improving perceptual coherence:

\begin{equation}
I'_{\text{fg}}(x,y) = \alpha(\Delta D(x,y)) \cdot I_{\text{fg}}(x,y) + \beta(\Delta D(x,y)),
\end{equation}
where the scaling functions are defined as:

\begin{equation}
\alpha(\Delta D) = 1 + \lambda_1 \cdot \Delta D, \quad \beta(\Delta D) = \lambda_2 \cdot \Delta D.
\end{equation}

Here, \( \lambda_1 \) and \( \lambda_2 \) are contrast and brightness adjustment factors, respectively, and control the degree of modulation based on the depth misalignment. In our implementation, we empirically set \( \lambda_1 = 0.5 \) and \( \lambda_2 = 0.3 \); further sensitivity analysis is provided in Appendix.

This depth-guided refinement effectively integrates the transformed foreground into the spatial and photometric context of the background, thereby reducing visual artifacts and improving realism under complex scene geometry and lighting.

\textbf{Final Composition.}  
After depth-aware refinement, the adjusted foreground \( I'_{\text{fg}} \) is integrated with the reconstructed background \( \hat{I}_{\text{bg}} \) using a simple per-pixel composition function \( \mathcal{C} \):
\begin{equation}
\begin{aligned}
\hat{I}(x, y) &= \mathcal{C}(\hat{I}_{\text{bg}}(x, y), I'_{\text{fg}}(x, y), M(x, y)) \\
&=
\begin{cases}
I'_{\text{fg}}(x, y), & \text{if } M(x, y) = 1 \\
\hat{I}_{\text{bg}}(x, y), & \text{otherwise}
\end{cases}
\end{aligned}
\end{equation}
where \( M \in \{0, 1\}^{H \times W} \) is the binary text mask. The function \( \mathcal{C} \) directly replaces the masked region with the adjusted foreground while preserving the background elsewhere. Though simple, this strategy benefits greatly from prior depth-aware adjustment, which ensures local consistency in illumination and geometry.

\begin{algorithm}[t]
\caption{\textit{DanceText} Framework}
\label{alg:dancetext}
\begin{algorithmic}[1]
\REQUIRE Input image $I \in \mathbb{R}^{H \times W \times 3}$, prompt set $\{S_i\}_{i=1}^N$
\ENSURE Final output image $\hat{I}$

\vspace{0.5em}
\textcolor{gray}{\# Stage 1: Foreground Extraction}
\STATE $\{(B_i, T_i)\}_{i=1}^N \gets \mathcal{D}(I)$ \hfill \textcolor{gray}{// OCR-based text detection and recognition}
\STATE $M \gets \textsc{GenerateMask}(\{B_i\})$ \hfill \textcolor{gray}{// Initial mask from bounding boxes}
\STATE $\hat{M} \gets \mathcal{S}_{\text{SAM}}(I, M)$ \hfill \textcolor{gray}{// Glyph segmentation via SAM}
\STATE $\hat{M} \gets \textsc{KMeansFilter}(\hat{M})$ \hfill \textcolor{gray}{// Cluster filtering to suppress background noise}
\STATE $I_{\text{fg}} \gets I \odot \hat{M}$ \hfill \textcolor{gray}{// Extract foreground layer}

\vspace{0.5em}
\textcolor{gray}{\# Stage 2: Background Restoration}
\STATE $\hat{I}_{\text{bg}} \gets \mathcal{L}(I, M)$ \hfill \textcolor{gray}{// Inpaint background via LaMa}

\vspace{0.5em}
\textcolor{gray}{\# Stage 3: Text Tampering and Transformation}
\FOR{$i = 1$ to $N$}
    \STATE $T'_i \gets \mathcal{T}(S_i, T_i)$ \hfill \textcolor{gray}{// Edit text using prompt-guided model}
    \STATE $B'_i \gets \mathcal{T}_{\text{geo}}(B_i, \text{params}_i)$ \hfill \textcolor{gray}{// Apply user-controlled geometric transformation}
\ENDFOR

\vspace{0.5em}
\textcolor{gray}{\# Stage 4: Depth-Aware Composition}
\STATE $D_{\text{bg}} \gets \textsc{EstimateDepth}(\hat{I}_{\text{bg}})$ \hfill \textcolor{gray}{// Estimate depth from inpainted background}
\STATE $D_{\text{fg}} \gets \textsc{EstimateDepth}(I_{\text{fg}})$ \hfill \textcolor{gray}{// Estimate depth of transformed foreground}
\STATE $\Delta D \gets D_{\text{bg}} - D_{\text{fg}}$ \hfill \textcolor{gray}{// Compute spatial depth mismatch}
\STATE $I'_{\text{fg}} \gets \mathcal{F}(I_{\text{fg}}, \Delta D)$ \hfill \textcolor{gray}{// Adjust brightness and contrast}
\STATE $\hat{I} \gets \mathcal{C}(\hat{I}_{\text{bg}}, I'_{\text{fg}}, \hat{M})$ \hfill \textcolor{gray}{// Final per-pixel composition}

\vspace{0.5em}
\RETURN $\hat{I}$
\end{algorithmic}
\end{algorithm}

\begin{figure*}[!ht]
    \centering
    \includegraphics[width=0.75\linewidth]{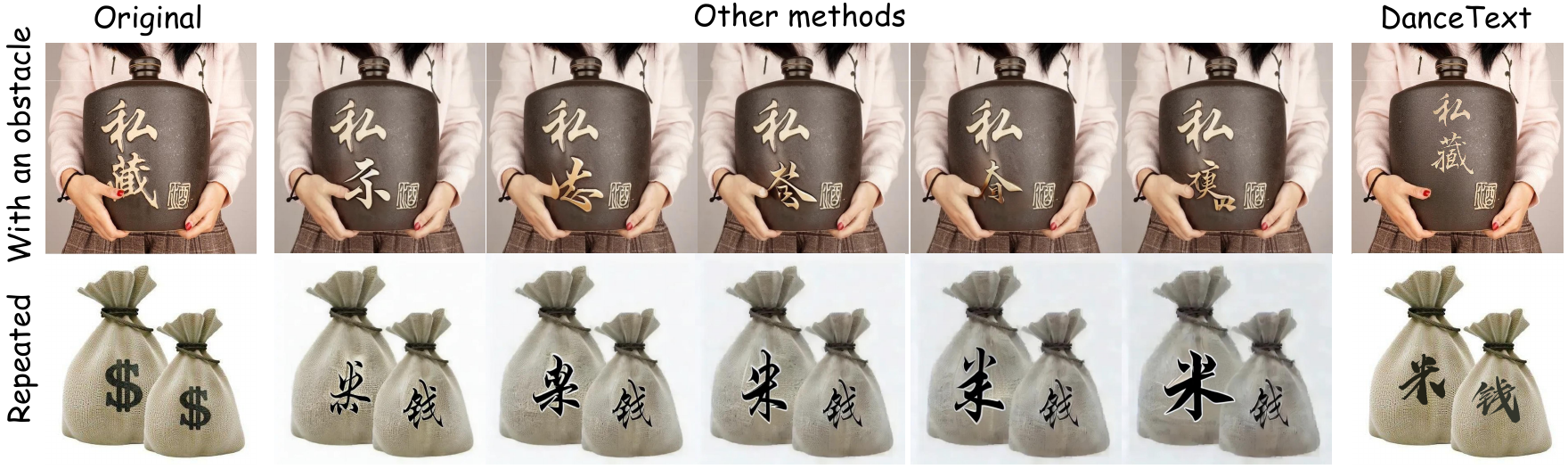}
    \caption{Merits and limitations of \textit{DanceText}, including its ability to edit occluded text, support multiple transformations, and preserve background consistency. Note: The images are sourced from the AnyWord-3M dataset.}
    \label{fig_merits}
\end{figure*}

\begin{figure*}[!ht]
    \centering
    \includegraphics[width=0.75\linewidth]{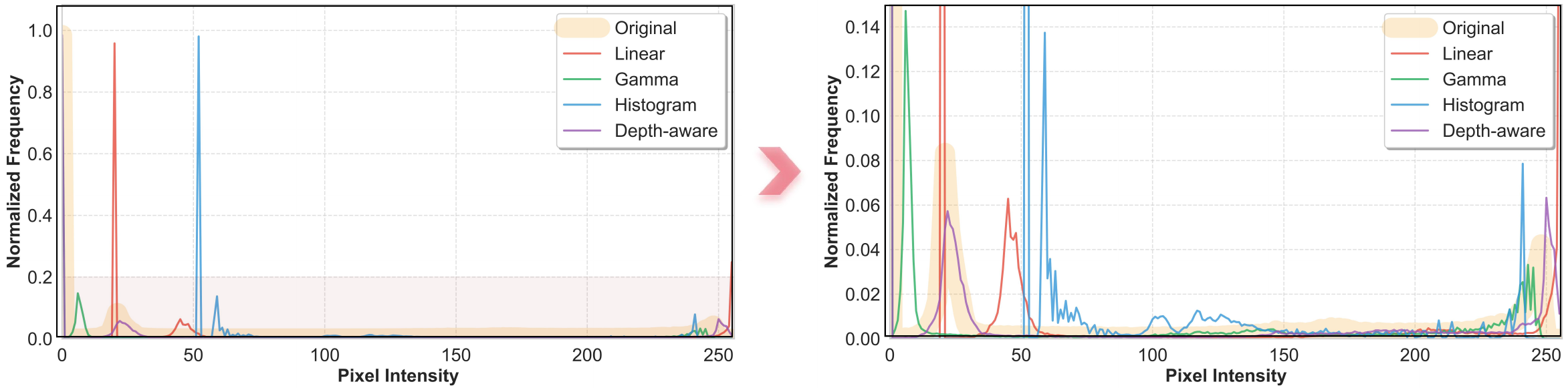}
    \caption{Pixel intensity histogram comparison of Linear, Gamma, Histogram matching, and Depth-aware adjustment methods.}
    \label{fig_ablation_histogram}
\end{figure*}

\section{Experiment}

\subsection{Dataset Description}

We use the OCR subset of the {AnyWord-3M} \cite{tuo2024anytext} dataset for experiments. This dataset is constructed from various publicly available image resources and encompasses a diverse range of multilingual text scenarios. The primary sources of images include {LAION-400M} \cite{schuhmann2021laion} and {Wukong} \cite{gu2022wukong}. Additionally, it integrates several benchmark datasets specifically designed for OCR recognition, such as COCO-Text \cite{veit2016coco}, RCTW \cite{shi2017icdar2017}, RRC-LSVT \cite{sun2019icdar}, MLT \cite{nayef2019icdar2019}, and ReCTS \cite{zhang2019icdar}.

\subsection{Experimental Settings}
All experiments were conducted on an NVIDIA GeForce RTX 4090 GPU with 24GB of memory. Since \textit{DanceText} does not require additional training, we ensured reproducibility by running all methods under the same hardware environment and software dependencies. For comparison, we selected models such as SAM, LaMa, and AnyText-v1.1, strictly using their officially released default parameters to ensure fairness in the evaluation.

\subsection{Evaluation Metrics}

\textbf{Text editing quality.}
    Sentence Accuracy (SA) \cite{tuo2024anytext} measures whether the OCR-recognized text matches the target text exactly, assessing the precision of text editing.
    Normalized Edit Distance (NED) \cite{tuo2024anytext} computes the similarity between the generated and target text; lower NED values indicate better editing performance.
    Frechet Inception Distance (FID) \cite{heusel2017gans} evaluates the visual style difference between the generated and real images; a lower FID indicates higher visual consistency.
During training, we use PP-OCRv3 \cite{li2022pp} for feature extraction. To ensure fairness in evaluation, we adopt DuGuangOCR \cite{duguangocr2023} as an independent OCR model for testing.

\textbf{Depth-Aware module.}
    Bhattacharyya Coefficient (BC) \cite{bhattacharyya1943measure} \& Correlation (Corr) \cite{pearson1895note} assess the consistency of depth features before and after text relocation. Higher values indicate stable visual depth preservation after text movement.
    Chi-Square Distance (CS) \cite{dodge2003oxford} \& Histogram Intersection (Inter) \cite{swain1991color} measure the impact of text relocation on depth information, ensuring smooth integration of the edited text with the background.

\subsection{Comparison}

\textbf{Comparison of generative methods}.  
We compare multiple generative approaches, including TextDiffuser, AnyText, Stable Diffusion V1.5/2.1/3.0/3.5, and DALL-E. As shown in Figure~A1 and ~A2, these models exhibit varying text rendering quality, visual coherence, and adaptability across different prompts. However, despite these variations, the choice of generative model does not significantly affect the effectiveness of our subsequent text editing and transformation steps. This suggests that \textit{DanceText} is robust to different input images, ensuring flexible and high-quality text manipulation regardless of the source image generation method.

\textbf{Comparison of depth estimation methods}.
Depth estimation plays a crucial role in ensuring natural foreground-background blending during text transformation. We compare multiple depth estimation models, including Marigold \cite{ke2023repurposing}, Geowizard \cite{fu2024geowizard}, and Depth Anything V2 (DAv2) \cite{depth_anything_v2}. As seen in Figure~A3, DAv2 produces the most stable and consistent depth maps, which are essential for our depth-aware fusion process. The smoothness of DAv2’s depth predictions helps maintain visual consistency when applying transformations like rotation, translation, resizing, and warping. Based on this evaluation, we selected DAv2 as the final depth estimation method in \textit{DanceText}, enhancing robustness across diverse scenarios.

\textbf{Comparison of different text transformations.}
Figure~\ref{fig_result_dancetext} illustrates the results of applying different transformation techniques, including rotation, translation, resizing, and warping. Among these, warping provides the highest flexibility, allowing text to conform to complex surfaces and non-linear distortions. However, wrap-based transformations can introduce artifacts if not properly blended with the background. To address this, \textit{DanceText} incorporates a depth-aware module that adjusts brightness and contrast based on local depth variations, ensuring seamless text-background integration. This enhancement minimizes visual artifacts and significantly improves the realism of the transformed text.

\textbf{Comparison of different editing models.}
Table~\ref{table_compare_different_methods} summarizes the performance of \textit{DanceText} and several state-of-the-art text editing baselines on the AnyWord-3M benchmark. In English tasks, \textit{DanceText} achieves a SA of 0.7011 and an NED of 0.8702, ranking second only to AnyText-v1.1 (SA = 0.7239, NED = 0.876), while outperforming all other methods. Notably, it achieves a lower FID (33.91) than AnyText-v1.0 (35.87), suggesting better visual realism.

In Chinese tasks, \textit{DanceText} achieves an NED of 0.8165 and an FID of 35.15, again placing it among the top three methods. Although AnyText-v1.1 achieves the best performance in most metrics, it is fine-tuned on the full benchmark, while \textit{DanceText} operates in a training-free fashion with no additional tuning. This highlights its competitive generalization ability.

Moreover, unlike methods such as AnyText that primarily support static or localized editing, \textit{DanceText} enables complex geometric transformations while maintaining robust text fidelity and perceptual quality. These results demonstrate the effectiveness of our layered architecture and depth-aware composition in handling transformations with minimal degradation.

\begin{table}[!ht]
    \centering
    \caption{Comparison on English and Chinese text editing tasks. All baseline results are reported from the AnyWord-3M benchmark. GlyphControl is fine-tuned on the TextCaps-5k dataset.}
    \resizebox{0.9\linewidth}{!}{
    \begin{tabular}{c|ccc|ccc}
    \toprule
        \multirow{2}{*}{\textbf{Method}} & \multicolumn{3}{c|}{\textbf{English}} & \multicolumn{3}{c}{\textbf{Chinese}} \\ \cline{2-7}
        ~ & \textbf{SA}$\uparrow$ & \textbf{NED}$\uparrow$ & \textbf{FID}$\downarrow$ & \textbf{SA}$\uparrow$ & \textbf{NED}$\uparrow$ & \textbf{FID}$\downarrow$ \\ \midrule
        ControlNet & 0.5837 & 0.8015 & 45.41 & 0.362 & 0.6227 & 41.86 \\ 
        TextDiffuser & 0.5921 & 0.7951 & 41.31 & 0.0605 & 0.1262 & 53.37 \\ 
        GlyphControl & 0.5262 & 0.7529 & 43.1 & 0.0454 & 0.1017 & 49.51 \\ 
        AnyText-v1.0 & \textit{0.6588} & \underline{0.8568} & \textit{35.87} & \underline{0.6634} & \underline{0.8264} & \textbf{28.46} \\ 
        AnyText-v1.1 & \textbf{0.7239} & \textbf{0.876} & \textbf{33.54} & \textbf{0.6923} & \textbf{0.8396} & \underline{31.58} \\ 
        \textit{DanceText} & \underline{0.7011} & \textit{0.8702} & \underline{33.91}  & \textit{0.6428} & \textit{0.8165} & \textit{35.15}  \\ \bottomrule
    \end{tabular}
    }
    \label{table_compare_different_methods}
\end{table}

\begin{table}[!ht]
    \centering
    \caption{Ablation study on the depth-aware module (D) for transformations in \textit{DanceText}. Each transformation is evaluated with and without depth-aware adjustment.}
    \resizebox{1.0\linewidth}{!}{
    \begin{tabular}{c|c|ccc|ccc}
    \toprule
    \multirow{2}{*}{\textbf{Transformation}} & \multirow{2}{*}{\textbf{D}} & \multicolumn{3}{c|}{\textbf{English}} & \multicolumn{3}{c}{\textbf{Chinese}} \\ \cline{3-8}
     & & \textbf{SA}$\uparrow$ & \textbf{NED}$\uparrow$ & \textbf{FID}$\downarrow$ & \textbf{SA}$\uparrow$ & \textbf{NED}$\uparrow$ & \textbf{FID}$\downarrow$ \\
    \midrule
    Rotation    & $\times$ & 0.6934 & 0.8655 & \underline{34.81} & \underline{0.6426} & 0.8135 & \underline{35.42} \\
    Translation & $\times$ & \textbf{0.7008} & \textbf{0.8679} & 34.87 & 0.6386 & \underline{0.8146} & 35.43 \\
    Resizing    & $\times$ & \underline{0.6997} & 0.8638 & \textbf{33.52} & \textbf{0.6465} & \textbf{0.8209} & \textbf{35.12} \\
    Wrap        & $\times$ & 0.6893 & \underline{0.8664} & 35.85 & 0.6334 & 0.8125 & 36.33 \\
    \midrule
    Rotation    & $\checkmark$ & 0.7010 & 0.8717 & 33.61 & 0.6477 & 0.8160 & 35.14 \\
    Translation & $\checkmark$ & \textbf{0.7075} & \textbf{0.8731} & \textbf{32.41} & \textbf{0.6539} & \textbf{0.8229} & \textbf{34.87} \\
    Resizing    & $\checkmark$ & \underline{0.7022} & \underline{0.8719} & \underline{32.45} & \underline{0.6482} & \underline{0.8224} & \underline{34.91} \\
    Wrap        & $\checkmark$ & 0.7008 & 0.8698 & 34.17 & 0.6362 & 0.8147 & 36.08 \\
    \bottomrule
    \end{tabular}
    }
    \label{table_compare_model_part}
\end{table}

\begin{table}[!ht]
    \centering
    \caption{Ablation study on the impact of the depth-aware module.}
    \resizebox{0.75\linewidth}{!}{
    \begin{tabular}{ccccc}
    \toprule
        \textbf{Method} & \textbf{BC}$\uparrow$ & \textbf{CS}$\downarrow$ & \textbf{Corr}$\uparrow$ & \textbf{Inter}$\uparrow$ \\ \midrule
        Linear & \underline{0.7882}  & 63.2225  & 0.0388  & 0.3965  \\ 
        Gamma & 0.4205  & \underline{15.2911}  & \underline{0.9649}  & \underline{1.5315}  \\ 
        Histogram & \textbf{0.8465}  & 64.5014  & 0.0141  & 0.3329  \\ 
        Depth-aware & 0.2546  & \textbf{0.7881}  & \textbf{0.9915}  & \textbf{1.8251}  \\ \bottomrule
    \end{tabular}
    }
    \label{table_ablation_depth_aware}
\end{table}

\subsection{Ablation Study}
\label{sec:ablation}
\textbf{Depth-aware adjustment on transformations.}
We evaluate four types of geometric transformations—rotation, translation, scaling, and warping—under two configurations: with and without the depth-aware adjustment module (Table~\ref{table_compare_model_part}). The depth-aware module consistently improves all evaluation metrics (SA, NED, and FID) across both English and Chinese text. For instance, in the translation scenario, the FID improves from 34.87 to 32.41 (English) and from 35.43 to 34.87 (Chinese), while NED increases from 0.8679 to 0.8731 (English), demonstrating more faithful text preservation and enhanced realism. Notably, the benefits of depth-aware adjustment are more pronounced in complex transformations such as warping, which tend to disrupt photometric consistency without such correction.

\textbf{Composition strategy on visual integration.}
We compare four foreground-background composition strategies: Linear, Gamma, Histogram matching, and our proposed Depth-aware method (Figure~\ref{fig_compare_composition}). Traditional methods often result in visual artifacts—e.g., over-saturation in linear scaling or hue shifts in histogram matching. In contrast, the depth-aware composition achieves smoother integration, maintaining local illumination consistency and reducing boundary discontinuities. These findings suggest that depth information plays a critical role in aligning text with complex visual scenes, especially under varying lighting or curved surfaces.

\textbf{Brightness and contrast adjustment methods.}
To further assess the effectiveness of our depth-aware adjustment, we conduct a pixel-wise histogram similarity analysis against the original background distribution. As shown in Figure~\ref{fig_ablation_histogram} and Table~\ref{table_ablation_depth_aware}, the depth-aware method achieves the highest correlation (0.9915) and intersection (1.8251), while also attaining the lowest Bhattacharyya coefficient (0.2546) and Chi-Square distance (0.7881). These results quantitatively confirm that our method preserves global tonal distribution better than baseline techniques, validating its perceptual superiority in text-background fusion.

\section{Merits and Limitations}

\subsection{Merits}
    \textbf{(1) Support for occluded text editing} (Figure~\ref{fig_merits}). Unlike traditional text editing methods, \textit{DanceText} effectively handles occluded text regions, enabling natural modifications without introducing noticeable artifacts. By leveraging foreground-background separation and the depth-aware module, the system maintains high visual consistency even in complex scenes.
    \textbf{(2) Support for repeated edits without affecting the background.} Due to the layered editing strategy, \textit{DanceText} allows multiple modifications to the same region without progressively degrading the background quality. This ensures that iterative edits retain high visual fidelity and prevent unwanted artifacts from accumulating. Further constraints can be easily added in our pipeline to adhere compliance with the best practices of responsible generative artificial intelligence~\cite{raza2025responsible}.
\subsection{Limitations}
    \textbf{Font and color consistency needs improvement.} While \textit{DanceText} effectively relocates and modifies text, slight inconsistencies in font style or color may appear in some cases, especially when blending with highly complex backgrounds. Future work could explore adaptive font rendering techniques to further enhance consistency.

\section{Conclusion}

We present \textit{DanceText}, a novel multilingual text editing framework that enables high-quality transformation, flexible editing, and seamless foreground-background integration. By adopting a {layered editing strategy}, \textit{DanceText} decouples text from background content, supporting user-controllable geometric transformations; including translation, rotation, scaling, and warping—while preserving spatial and photometric consistency. The framework integrates OCR and SAM for foreground extraction, LaMa-based inpainting for background restoration, and a {depth-aware adjustment module} for contrast and brightness alignment. Extensive experiments demonstrate that \textit{DanceText} achieves competitive visual quality and text recognition accuracy, particularly under large-scale and complex transformations. The proposed framework is highly applicable to real-world multimedia scenarios and can be extended to more complex transformations and animations.

\section*{Appendix}
The appendix can be downloaded from 
\href{https://pan.baidu.com/s/1cvFxYWH_EtMQJOcPOMwobg?pwd=yoyo}{here}.

\bibliographystyle{ieeetr}
\bibliography{main}

\vfill

\end{document}